\documentclass[runningheads]{llncs}

\PassOptionsToPackage{hyperfootnotes=false}{hyperref}

\usepackage{eccv}

\usepackage{eccvabbrv}

\usepackage{graphicx}
\usepackage{booktabs}
\usepackage{wrapfig}
\usepackage{marvosym} 
\usepackage[accsupp]{axessibility}  % Improves PDF readability for those with disabilities.

\usepackage{hyperref}

\usepackage{orcidlink}
\usepackage{multirow}
\usepackage{adjustbox}
\begin{document}

% ---------------------------------------------------------------
% TODO REVIEW: Replace with your title
\title{Unison: Harmonizing Motion, Speech, and Sound for Human-Centric Audio-Video Generation} 

% TODO REVIEW: If the paper title is too long for the running head, you can set
% an abbreviated paper title here. If not, comment out.
\titlerunning{Unison}

% TODO FINAL: Replace with your author list. 
% Include the authors' OCRID for the camera-ready version, if at all possible.
\author{Shihao Cheng\inst{1,4\dagger} \and
Jiaxu Zhang\inst{2\dagger} \and
Quanyue Song\inst{3,4} \and
Shansong Liu\inst{4\ddagger} \and
Zhizhi Guo\inst{4} \and
Xiao-Lei Zhang\inst{5} \and
Chi Zhang\inst{5} \and
Xuelong Li\inst{5} \and
Zhigang Tu\inst{1}\textsuperscript{\Letter}
}

% TODO FINAL: Replace with an abbreviated list of authors.
\authorrunning{S.~Cheng et al.}
% First names are abbreviated in the running head.
% If there are more than two authors, 'et al.' is used.

% TODO FINAL: Replace with your institution list.
\institute{State Key Laboratory of Information Engineering in Surveying, Mapping and Remote Sensing, Wuhan University, Wuhan, China \and
ByteDance, China \and
State Key Laboratory of Human-Machine Hybrid Augmented Intelligence, Institute of Artificial Intelligence and Robotics, Xi'an Jiaotong University, China \and
China Telecom Artificial Intelligence Technology (Beijing) Co., Ltd., China \and
Institute of Artificial Intelligence (TeleAI), China Telecom, China
}

\maketitle
\begingroup
\renewcommand\thefootnote{}\footnotetext{$^{\dagger}$ Equal contribution.}
\renewcommand\thefootnote{}\footnotetext{$^{\ddagger}$ Project leader.}
\renewcommand\thefootnote{}\footnotetext{\textsuperscript{\Letter} Corresponding author.}
\endgroup

\vspace{-17pt}
\begin{figure}[!h]
  \setlength{\abovecaptionskip}{-.2cm}
  \begin{center}
    \includegraphics[width=\linewidth]{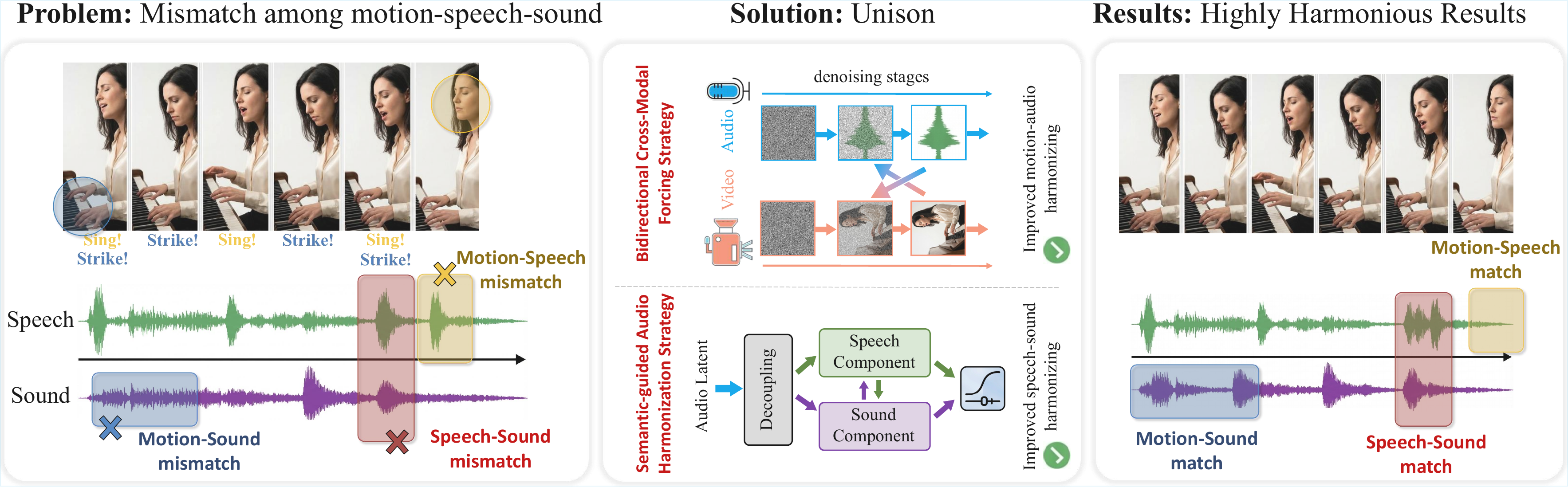}
  \end{center}
  \caption{\textbf{Overview of the key challenges and our approach.} \textbf{Left:} Two major misalignments in human-centric audio-video generation---(a) speech-sound interference within the audio stream and (b) motion-audio desynchronization, including motion-speech and motion-sound mismatches. \textbf{Middle:} Our proposed Unison framework addresses these issues through a semantic-guided harmonization strategy that decouples and adaptively balances speech-sound components via interactive refinement, and a progressive audio-video forcing strategy that enforces rigorous temporal alignment across modalities. \textbf{Right:} These designs jointly reduce speech dominance, preserve sound effects, and achieve coherent synchronization between motion and audio.}
  \label{fig:1}
\end{figure}
\vspace{-32pt}
\begin{abstract}
  Motion, speech, and sound effects are fundamental elements of human‑centric videos, yet their heterogeneous temporal characteristics make joint generation highly challenging. Existing audio‑video generation models often fail to maintain consistent alignment across these modalities, leading to noticeable mismatches between motion, speech, and environmental sounds.
  %
  % We present Unison, a unified framework that explicitly promotes coherence within and across the motion–speech–sound space. Within the audio branch, Unison introduces a bidirectional audio cross-attention that separates and adaptively recomposes speech and sound representations in band‑structured subspaces, enabling more precise control of their spectral interplay.
  We present Unison, a unified framework that explicitly promotes coherence across the motion, speech, and sound modalities.
  Within the audio stream, Unison employs a semantic-guided harmonization strategy that decouples the generation of speech and sound-effect components. Leveraging bidirectional audio cross-attention and semantic-conditioned gating for semantic-driven adaptive recomposition, this approach effectively mitigates speech dominance and enhances acoustic clarity.
  For audio–motion synchronization, we propose a bidirectional cross-modal forcing strategy where the cleaner modality guides the noisier one through decoupled denoising schedules, reinforced by a progressive stabilization strategy.
  Extensive experiments demonstrate that Unison achieves state‑of‑the‑art performance in both audio perceptual quality and cross‑modal synchronization, highlighting the importance of explicit multimodal harmonization in human‑centric video generation.
\end{abstract}

\section{Introduction}
\label{sec:intro}
The rapid evolution of generative AI~\cite{an2025aiflowperspectivesscenarios,11027794,wan2025wan,low2025ovi,10003114,chen2025clusterstyle,hacohen2026ltx2efficientjointaudiovisual,lipman2023flowmatchinggenerativemodeling,shen2025survey,li2024survey,zhang2024semtalk,zhang2025echomaskspeechqueriedattentionbasedmask,zhang2025performing,zhang2026leveraging,zhang2025visual,zhang2026dart,zhang2026mitigating,zhang2026personagesture,wang2026udmgrpostableefficientgroup,Xiao_2026_CVPR} has catalyzed significant breakthroughs in synchronized audio-video generation. Proprietary models such as Veo~3~\cite{deepmind2025veo}, Sora~2~\cite{openai2025sora2}, and Wan~2.5~\cite{wan2025wan}, alongside open-source models like Ovi~\cite{low2025ovi}, UniAVGen~\cite{zhang2025uniavgen}, and LTX-2~\cite{hacohen2026ltx2efficientjointaudiovisual}, have demonstrated remarkable creative potential, inspiring users across online communities. 

Although most open-source approaches have continuously improved visual fidelity~\cite{wang2025universe,jiang2024data}, they still exhibit poor alignment between audio and video streams. Recent methods, such as Harmony~\cite{hu2025harmony}, employ fine-grained cross-attention mechanisms to enable frame-level synchronization. However, these approaches primarily focus on architectural modifications and lack explicit consistency constraints. In contrast, closed-source commercial systems leverage large-scale datasets to achieve fully data-driven synchronization~\cite{openai2025sora2}, facilitating visually and temporally coherent outputs. Nevertheless, the generated audio often lacks acoustic richness and balance: speech components tend to dominate the entire mix, suppressing ambient and contextual sounds that are essential for perceptual realism. This imbalance leads to flattened auditory scenes and weakens the overall sense of immersion, with these issues becoming especially evident in human-centric video generation.

As shown in the left part of Figure~\ref{fig:1}, our evaluation of representative audio–visual generation models reveals two fundamental challenges. (1) Intra‑audio interference: speech and non‑speech sound effects often overlap within the same audio stream, creating perceptual masking where structured speech components dominate and obscure transient environmental sounds—especially in human‑centric scenarios such as ``singing while playing an instrument." This indicates that current models lack mechanisms to disentangle and balance heterogeneous acoustic elements. (2) Cross‑modal misalignment: motion and audio generation are typically linked only through architectural cross‑attention, without explicit synchronization objectives during training. Consequently, audio and video exchange information implicitly but fail to form consistent temporal correspondence, leading to desynchronized gestures, lip motions, and acoustic onsets. These issues collectively limit the realism and coherence of human‑centered video generation.

To overcome the aforementioned challenges, we propose Unison, a unified framework designed to generate motion, speech, and sound in a coherent and synchronized manner. 
To resolve acoustic interference, Unison implements a semantic-guided harmonization strategy that decouples speech and sound-effect generation. This approach integrates bidirectional audio cross-attention (Bi-ACA) for interactive refinement and semantic-conditioned gating (SCG) for semantic-driven adaptive recomposition. By modulating component interplay, Unison ensures cross-modal consistency while enabling dedicated modeling for each acoustic entity, effectively preventing speech from overshadowing subtle environmental sounds.
Second, to tackle motion--audio desynchronization, Unison employs a bidirectional cross-modal forcing paradigm that explicitly aligns denoising progress between modalities. By allowing the temporally advanced modality to guide the lagging one during training, the model learns stable and causally consistent motion--audio coordination without relying solely on cross-attention. Together, these designs establish a principled way of generating human-centric videos where motion, speech, and acoustic context evolve in Unison--yielding perceptually balanced and temporally synchronized results.

Extensive experiments verify that Unison not only enhances overall generation fidelity but also achieves markedly better harmony among motion, speech, and sound. The resulting videos exhibit more balanced auditory layers and tighter temporal correspondence between modalities, confirming that both spectral disentanglement and cross‑modal alignment substantially improve perceptual realism. Moreover, the framework’s symmetric design naturally enables bidirectional generation—it can synthesize video conditioned on audio or, conversely, generate audio from visual cues—demonstrating its flexibility as a unified model for multimodal content creation.

The main contributions are summarized as follows:

\begin{itemize}

\item We propose Unison, a unified audio–video generation framework that produces motion, speech, and sound in a coherent and harmonized manner. Unlike prior models that loosely couple audio and visual streams, Unison explicitly enforces cross‑modal consistency, achieving balanced and temporally aligned outputs.

\item We propose a semantic-guided harmonization strategy that decouples speech and sound-effect generation to resolve acoustic interference. By integrating Bidirectional Audio Cross-Attention (Bi-ACA) for interactive refinement and Semantic-Conditioned Gating (SCG) for semantic-driven adaptive recomposition, resulting in controllable and acoustically layered audio generation.

\item We introduce a bidirectional cross-modal forcing strategy that strengthens temporal dependencies between audio and motion. By allowing the modality in a clearer denoising state to guide the other, Unison achieves robust synchronization and stable training, effectively bridging the gap between motion and acoustics in human‑centric video generation.
\end{itemize}

\section{Related Work}

\subsection{Joint Audio-Video Generation}
Early joint audio-video generation, exemplified by MM-Diffusion~\cite{ruan2022mmdiffusion}, utilized U-Net architectures for ambient sound synthesis, a scope later expanded by large-scale dataset training~\cite{liu2024syncflowtemporallyalignedjoint, zhao2025uniformunifiedmultitaskdiffusion}. While UniVerse-1~\cite{wang2025universe} and Ovi~\cite{low2025ovi} recently introduced human speech generation, their reliance on implicit global alignment often compromises temporal precision. Advanced mechanisms~\cite{hu2025harmony, zhang2025uniavgen, hacohen2026ltx2efficientjointaudiovisual} have since achieved frame-level synchronization via fine-grained cross-attention. Nonetheless, these methods typically employ unidirectional conditioning, leaving bidirectional cross-modal reinforcement and the acoustic interference between speech and effects largely unaddressed.

Recent commercial systems, including Veo 3~\cite{deepmind2025veo}, Sora 2~\cite{openai2025sora2}, and Wan 2.5~\cite{wan2025wan}, have achieved cinematic audio-video synchronization. 
Nevertheless, when synthesizing scenes with concurrent dense speech and sound effects, these models often prioritize human voices, leading to the suppression or overriding of essential environmental acoustics. 
To address this, Unison implements a semantic-guided harmonization strategy that decouples speech and sound-effect generation. By isolating these acoustic components at the generative source, this approach resolves speech dominance and ensures intrinsic acoustic clarity.

\subsection{Diffusion Forcing}
Originally pioneered by Chen~\cite{chen2024diffusionforcingnexttokenprediction} to enable sequential modeling with independent per-step noise levels, the diffusion forcing paradigm has been extended to transformer-based architectures~\cite{song2025historyguidedvideodiffusion}, self-corrective long-term synthesis~\cite{huang2025selfforcingbridgingtraintest}, and sliding-window video generation~\cite{liu2025rollingforcingautoregressivelong,song2026interactiveavatarrealtimestreamingvideo}. While these advancements effectively mitigate exposure bias and background forgetting in unimodal sequences, they lack a systematic mechanism to capture bidirectional audio--visual dependencies across heterogeneous denoising processes. Unison generalizes this principle to a multimodal setting by introducing a cross-modal forcing strategy. This approach enables audio and video signals to mutually guide their respective denoising trajectories, thereby reinforcing cross-modal interdependencies for superior audiovisual synchronization.

\begin{figure*}[t]
\vspace{-.1cm}
\setlength{\abovecaptionskip}{-.1cm}
\setlength{\belowcaptionskip}{-.4cm}
  \begin{center}
      \includegraphics[width=\textwidth]{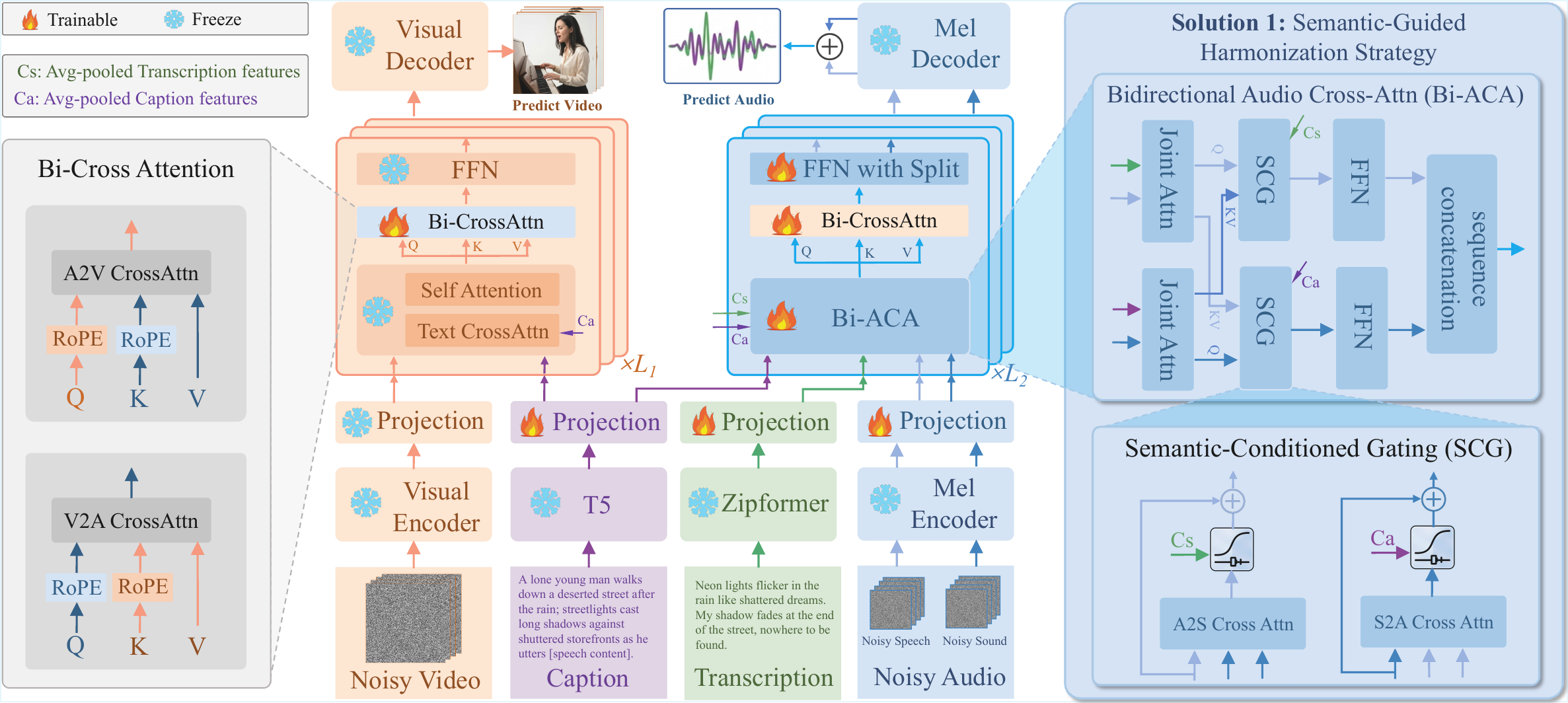}
  \end{center}
  \caption{\textbf{Overview of Unison.} Unison couples a video branch and an audio branch via bidirectional cross-attention. The audio branch employs a Semantic-Guided Harmonization Strategy for independent speech and sound-effect generation, utilizing a Bidirectional Audio Cross-Attention (Bi-ACA) module to mutually refine speech and sound-effect features, effectively enhancing their respective clarity. At each interaction node, a Semantic-Conditioned Gating (SCG) mechanism is employed that balances speech and sound-effect contributions based on $c_s$ and $c_a$.}
  \label{fig:overview}
\end{figure*}

\section{Method}
We introduce Unison, a unified audio-video generation framework designed to transform text prompts into synchronized audiovisual content that harmonizes motion, speech, and sound effects. Figure~\ref{fig:overview} illustrates our dual-branch architecture.
We formulate Unison's processes as follows. 
\begin{equation}
    \text{Unison}(\kappa, \tau, c_s, c_a) \mapsto (\nu, \alpha).
\end{equation}
where $\nu$ and $\alpha$ denote the output video and audio, respectively. The inputs consist of the caption text $\kappa$, the transcription text $\tau$, and their corresponding average-pooled feature representations, $c_a$ and $c_s$.

Unison employs a dual-branch architecture comprising a video branch based on Wan2.2-5B \cite{wan2025wan} and an audio branch built upon MMAudio \cite{cheng2025taming}, with the latter augmented by Zipformer \cite{zhu2025zipvoice,yao2024zipformerfasterbetterencoder} to enable speech generation. The video branch comprises 29 layers ($L_1{=}29$) and the audio branch 23 layers ($L_2{=}23$). Integration between these branches is achieved through frame-level bidirectional cross-attention, utilizing video and audio latents as mutual queries to facilitate seamless cross-modal information exchange.

To achieve word-level lip-sync, we enforce synchronization across three dimensions. At the architecture level, cross-attention aligns features within a three-frame window using a stride of one, and only the middle frame's representation is retained during restoration. At the data level, we introduce a lip-filtering operator that first detects the number and location of faces. SyncNet~\cite{Chung16a} is then applied exclusively within these bounding boxes to verify alignment, which naturally filtering out clips with unsynchronized speech and lip movements, as well as off-screen voice-overs. Finally, we design a Bidirectional Cross-Modal Forcing Strategy to enhance audio-visual synchronization by reinforcing the mutual dependence between the two modalities during training, as detailed in Section~\ref{subsec:motion_audio_coordination}.

\subsection{Preliminaries}

\paragraph{Flow Matching (FM)}~\cite{lipman2023flowmatchinggenerativemodeling} provides a simulation-free framework for training continuous normalizing flows by regressing a velocity field $v_\theta(x_t, t)$ onto a target vector field $u_t(x)$ that transports samples from a prior $p_0$ to the data distribution $p_1$. For a conditional probability path $p_t(x|x_1)$ with optimal transport, $x_t = (1{-}t)x_0 + t x_1$ and $u_t(x_t|x_1) = x_1 - x_0$. The Conditional Flow Matching (CFM) objective is:
\begin{equation}
    \mathcal{L}_{\text{CFM}} = \mathbb{E}_{t, x_1, x_t}\big[\|v_\theta(x_t, t) - (x_1 - x_0)\|^2\big].
\end{equation}
By defining a deterministic mapping via an associated ODE, FM offers straighter sampling trajectories and improved inference efficiency with fewer function evaluations compared to score-based diffusion.

\paragraph{Diffusion Forcing (DF)}~\cite{chen2024diffusionforcingnexttokenprediction} extends diffusion to sequences by assigning an \emph{independent} noise level to each element. For a sequence $\mathbf{x}_{1:L}$, it samples per-element timesteps $\mathbf{t}=(t_1,\dots,t_L)$, training the model with a heterogeneous-noise objective:
\begin{equation}
    \mathcal{L}_{\text{DF}} = \mathbb{E}_{\mathbf{x}, \mathbf{t}} \Big[\sum_{i=1}^{L}\big\|\mathcal{T}(x_i,t_i) - f_\theta(\mathbf{x}_{1:L,\mathbf{t}}, \mathbf{t})_i\big\|^2\Big],
\end{equation}
where $\mathcal{T}(x_i,t_i)$ is the target and $f_\theta(\cdot)_i$ denotes the prediction for element $i$ conditioned on the full heterogeneous sequence. As a diffusion analogue of next-token prediction, DF allows for flexible causal or bidirectional context windows, effectively reducing train--test mismatch and enhancing long-horizon stability by mitigating error accumulation.

\subsection{Speech-Sound Coordination}
We extend MMAudio \cite{cheng2025taming} by integrating Zipformer \cite{zhu2025zipvoice,yao2024zipformerfasterbetterencoder} to empower the model with robust speech generation capabilities. Reflecting the inherent structural divergence between speech and sound effects, we propose a Semantic-Guided Harmonization Strategy that decouples the generation process into separate speech and sound-effect streams to ensure high-fidelity synthesis and eliminate acoustic ambiguity. Under this strategy, a Bidirectional Audio Cross-Attention (Bi-ACA) module facilitates hierarchical feature interaction between these disentangled components, while a Semantic-Conditioned Gating (SCG) mechanism adaptively modulates their synthesis based on captions and transcriptions.

\paragraph{Bidirectional Audio Cross-Attention (Bi-ACA).} 
To enforce acoustic coherence while maintaining structural segregation, the audio latent is represented as a dual-stream tensor $h \in \mathbb{R}^{B \times 2 \times N \times D}$, comprising speech and sound-effect components. The source audio is decoupled and encoded into two temporally aligned sequences, $h^{sp}$ and $h^{sfx}$. Within each Transformer block, Bi-ACA facilitates mutual context exchange via bidirectional cross-attention before merging them for synchronized global modeling. Specifically, the latents are concatenated along the sequence dimension:
\begin{equation}
    h_{\text{joint}} = \text{Concat}([\tilde{h}^{sp}, \tilde{h}^{sfx}], \text{dim}=N) \in \mathbb{R}^{B \times 2N \times D}.
\end{equation}
To ensure precise temporal synchronization, both streams reuse identical temporal indices for Rotary Positional Embeddings (RoPE). To distinguish between the two modalities within the shared self-attention layers, we incorporate a modality-specific learnable bias, preventing semantic confusion while allowing for joint acoustic modeling. After processing $h_{\text{joint}}$ through the shared Transformer backbone, the representation is bifurcated at the block's exit to recover the independent generation trajectories:
\begin{equation}
    h^{sp}, h^{sfx} = \text{Split}(h_{\text{joint}}, \text{dim}=N).
\end{equation}
This interact-merge-split cycle enables the two streams to benefit from shared global context while maintaining their unique structural characteristics.

Bi-ACA executes \emph{bidirectional} cross-attention between the two streams across multiple decoder layers (Fig.~\ref{fig:overview}, right), facilitating intra-audio context exchange while preserving their structural independence:
\begin{equation}
    \tilde{h}^{[sp \mid sfx]} = h^{[sp \mid sfx]} + \text{Attn}(\text{LN}(h^{[sp \mid sfx]}), \text{LN}(h^{[sfx \mid sp]}))
\end{equation}
where $\mathrm{Attn}(Q, K, V)$ follows the standard multi-head cross-attention mechanism and $\mathrm{LN}$ denotes layer normalization. While individual speech or sound-effect streams may lack sufficient acoustic grounding when performing cross-attention with video tokens, Bi-ACA provides mutual audio-to-audio priors. This ensures that the synthesized components are not only visually aligned but also acoustically consistent, preventing the high-fidelity synthesis from degrading due to isolated feature mapping.

\paragraph{Semantic-Conditioned Gating (SCG).}
To preclude acoustic interference and phonetic degradation from unconstrained context injection, SCG adaptively regulates cross-stream updates via modality-specific semantic priors. Global semantic vectors $c_s$ and $c_a$ are derived through average pooling of transcription and caption sequences to predict dual gating coefficients $[g^{sp}, g^{sfx}] = \sigma(\text{MLP}([c_s; c_a]))$. These coefficients function as semantic filters to prioritize salient content. In narration-dominant scenes, SCG attenuates the influx of sound-effect features to safeguard phonetic purity. The mechanism further modulates the gating to accommodate intricate environmental layers in complex soundscapes like musical performances. The gated updates are formulated as:
\begin{equation}
    h^{[sp \mid sfx]}_{\text{out}} = h^{[sp \mid sfx]} + g^{[sp \mid sfx]} \cdot \text{Attn}(\text{LN}(h^{[sp \mid sfx]}), \text{LN}(h^{[sfx \mid sp]}))
\end{equation}
The coefficients $g \in [0, 1]$, constrained by the sigmoid function, act as semantic valves to regulate information flow. For instances with prominent $c_s$ (e.g., dense narration), SCG adaptively suppresses $g^{sp}$ to shield the speech stream from SFX interference, preserving phonetic purity. Conversely, in scenarios dominated by $c_a$ (e.g., complex soundscapes), it amplifies cross-stream influence to enrich non-speech acoustics, thereby achieving a balanced and controllable audio recomposition.

\paragraph{Audio Stream Supervision.}
To enforce explicit supervision within the decoupled architecture, we leverage Mel-Roformer to disentangle mixed audio into high-fidelity speech and sound-effect (SFX) components, which are encoded as ground-truth latents $z^{\text{sp}}_1$ and $z^{\text{sfx}}_1$. During training, the dual-stream decoder is optimized via independent flow-matching losses: $\mathcal{L}_{\text{dual}} = \mathcal{L}_{\text{CFM}}^{\text{sp}} + \mathcal{L}_{\text{CFM}}^{\text{sfx}}$. This explicit per-stream supervision eliminates acoustic ambiguity and source interference, ensuring that the decoupled representations remain independent and precisely aligned with the visual modality.

\subsection{Motion-Audio Coordination}
\label{subsec:motion_audio_coordination}
Regarding the structural design of audio-visual alignment, we follow established methods~\cite{hu2025harmony, zhang2025uniavgen} and employ frame-level cross-attention as illustrated in the left part of Figure~\ref{fig:overview}. Regarding the training strategy for audio-visual alignment, unlike prior works~\cite{hu2025harmony, zhang2025uniavgen} using identical timesteps, we independently sample timesteps for each modality. This allows the "cleaner" modality to guide the "noisier" one, forcing the model to rely more heavily on cross-modal information. Consequently, this bidirectional guidance strengthens the mutual dependency between audio and video.

\begin{figure*}[t] % 规范推荐使用 [t] 避开 [h]
  \vspace{-.1cm}
  \setlength{\abovecaptionskip}{-.1cm}
  \setlength{\belowcaptionskip}{-.4cm}
  \begin{center}      
    \includegraphics[width=\textwidth]{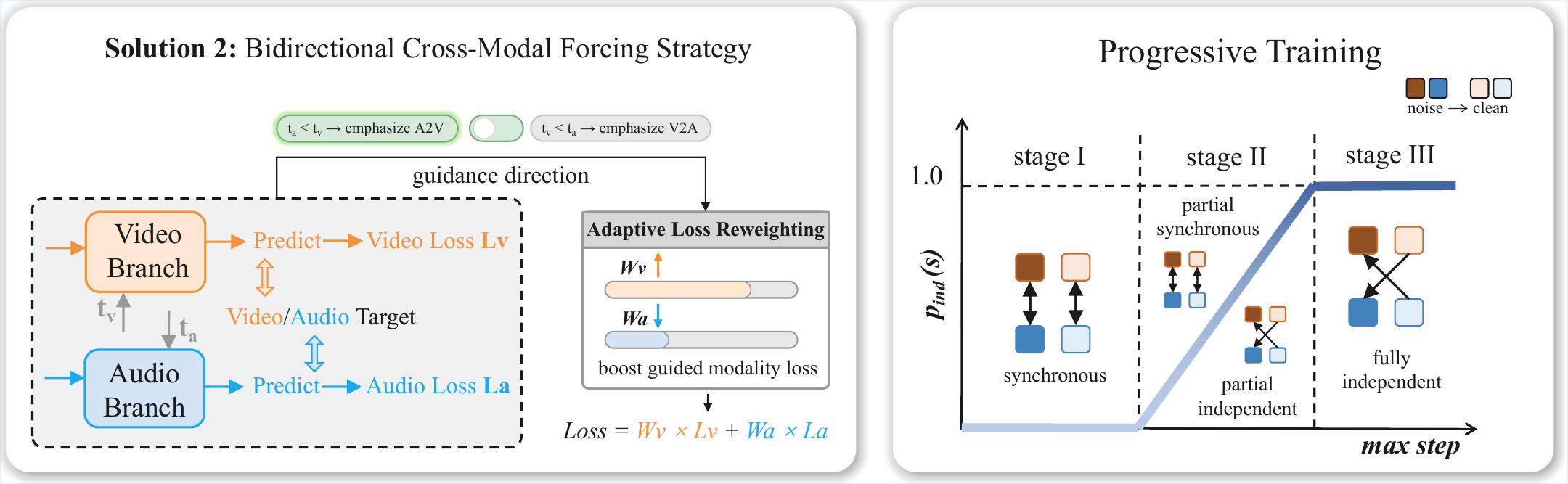}
  \end{center}
  \caption{\textbf{Bidirectional Cross-Modal Forcing strategy for audio-visual alignment.} By decoupling diffusion timesteps during training, our approach enables mutual guidance where the modality at a lower noise level provides enhanced conditioning to steer the denoising process of the noisier counterpart. To ensure stable optimization, a three-stage curriculum is employed, progressing from synchronous warmup to partial and eventually full temporal independence.}
  \label{fig:method_df}
\end{figure*} 

\paragraph{Bidirectional Cross-Modal Forcing.}
Inspired by \emph{Diffusion Forcing}~\cite{chen2024diffusionforcingnexttokenprediction}, we decouple the denoising timesteps for video and audio, allowing each modality to evolve through independent noise levels rather than being strictly synchronized. Specifically, we separately sample $t_v$ and $t_a$ per iteration to generate their respective noisy latents, with the audio branch mapped to the $[0, 1]$ interval. This asynchronous schedule enables the model to learn cross-modal dependencies across varying corruption levels.

Our core premise is that smaller timesteps yield \emph{cleaner} states with superior semantic reliability. Accordingly, we define a per-sample direction indicator $d = \mathbb{I}[t_a < t_v]$, where $d=1$ (or $d=0$) identifies audio (or video) as the leading modality. Rather than altering the fusion mechanism itself, $d$ dynamically designates the \emph{student} modality under noise-mismatched conditions. By upweighting the loss for the noisier branch, we compel it to extract stronger supervisory signals from the cleaner counterpart via cross-modal conditioning.

To prevent optimization instability from extreme noise gaps, we constrain $|t_v - t_a| \leq \Delta_{\max}$ (with $\Delta_{\max}=0.25$) and compute direction-aware loss weights governed by a guidance strength $\lambda=0.5$:
\begin{equation}
    w_v = 1 + \lambda \cdot d, \quad w_a = 1 + \lambda \cdot (1-d).
\end{equation}
The full TI2AV (Text-and-Image-to-Audio-Video) training objective is then defined as the direction-weighted sum of per-branch flow-matching losses:
\begin{equation}
    \mathcal{L}_{\text{TI2AV}} = w_v \cdot \mathcal{L}_{\text{CFM}}^{v} + w_a \cdot (\mathcal{L}_{\text{CFM}}^{\text{sp}} + \mathcal{L}_{\text{CFM}}^{\text{sfx}}),
    \label{eq:ti2av}
\end{equation}
where $\mathcal{L}_{\text{CFM}}^{v}$ is the video flow-matching loss, and $\mathcal{L}_{\text{CFM}}^{\text{sp}}$, $\mathcal{L}_{\text{CFM}}^{\text{sfx}}$ are the per-stream audio losses from Sec.~3.2. This strategy prioritizes the learning signal for the more heavily corrupted modality, significantly enhancing motion-audio correspondence across disparate convergence rates.

\paragraph{Progressive Training Strategy.}
Directly optimizing with fully independent timesteps precipitates training instability due to pronounced cross-modal noise disparities and stochastic oscillations in the leading direction $d$. We mitigate this via a three-stage curriculum: (i) Synchronous Warmup, enforcing $t_v = t_a$ to establish foundational alignment; (ii) Incremental Decoupling, activating independent sampling with probability $p_{\text{ind}}(s)$ while constraining $|t_v - t_a| \leq 0.25$; and (iii) Full Independence, employing unconstrained decoupled updates. To stabilize this manifold, we apply direction-aware loss reweighting starting from Phase II, up-scaling the video loss for $d=1$ (cleaner audio) and prioritizing the audio loss for $d=0$ (cleaner video). This progression ensures a robust optimization trajectory by gradually introducing increasingly complex cross-modal noise configurations.

\section{Experiments}
\subsection{Experimental Setup}
\paragraph{Training Corpora.} 
Our training framework leverages a diverse collection of audio-visual and audio-only corpora, incorporating Mel-Roformer~\cite{wang2024melroformervocalseparationvocal} to decouple both audio-visual tracks and mixed audio streams into independent speech and sound-effect components. For joint audio-visual training, we aggregate several large-scale open-source datasets including OpenHumanVid~\cite{li2025openhumanvidlargescalehighqualitydataset}, HDTF~\cite{zhang2021flow}, VFHQ~\cite{wang2022vfhqhighqualitydatasetbenchmark}, CelebV-Text~\cite{yu2022celebvtext}, and VGGSound~\cite{Chen20}. Regarding the audio branch training, we curate a comprehensive repository spanning speech, sound effects, music, and singing performances. Sound effects are collected from YouTube-8M~\cite{45619}, AudioSet~\cite{jort_audioset_2017}, and WavCaps~\cite{mei2023wavcaps}. Music data is derived from VidMuse~\cite{tian2025vidmuse}, and the singing portion is primarily extracted from the Yue collection~\cite{yuan2025yue, yuan2025yuescalingopenfoundation}. We also include internal speech data to further enrich the diversity and coverage of the training corpus. After refinement through our automated processing pipeline, the final dataset encompasses approximately 2 million synchronized audio-visual clips totaling over 3,000 hours, alongside 50 million high-quality audio segments exceeding 130,000 hours in total duration.

% \paragraph{Data Processing}
% To further enhance data quality, we developed a pipeline consisting of multiple filters to evaluate samples across three dimensions: video quality, audio quality, and audio-visual alignment, followed by labeling the high-quality data. Regarding video quality, we first exclude videos with aspect ratios greater than 2 or less than 0.5, then perform hard-cut detection on subtitled data, followed by spatial resolution filtering and the removal of videos with severe motion blur. For audio quality, we perform audio separation and Automatic Speech Recognition (ASR) on all videos, followed by audio integrity detection and temporal segmentation. To detect audio-visual alignment, we adopt a SyncNet-based approach involving S3FD face detection, tracking, feature extraction, and distance calculation, finally applying threshold filtering based on LSE-C and LSE-D; detailed specifications are provided in the Appendix.

\paragraph{Implementation Details.} 
Unison training involves two stages. Stage 1 (Audio Branch) utilizes 4 NVIDIA H100 GPUs with a 96-sample batch size. We adopt a $1\times 10^{-4}$ learning rate with 1k-step linear warmup and step decay ($\gamma=0.1$) at 240k and 270k steps. Stage 2 (Joint Training) fine-tunes the coupled system via the TI2AV objective (Eq.~\ref{eq:ti2av}) on 16 H100 GPUs using bf16 precision and ZeRO-2 optimization. At a $2\times 10^{-5}$ learning rate and 32-sample batch size, we optimize only the audio branch and fusion modules (bi-directional cross-attention and layer normalization), keeping the video backbone frozen. A progressive training strategy with phase ratios of 0.3, 0.4, and 0.3 ensures multimodal alignment stability. Inference employs a 50-step flow-matching sampler with classifier-free guidance (scale=6.0), producing 25 FPS videos. The code and models will be made publicly available upon acceptance.

\paragraph{Evaluation Metrics.} 
We report a comprehensive set of objective metrics. (1) For video assessment, we employ LAION-Aesthetic Predictor V2.5~\cite{schuhmann2022laion} to compute the Video Aesthetic Score (VA) as a proxy for artistic coherence, alongside DINOv3~\cite{simeoni2025dinov3} to measure inter-frame Identity consistency (ID). (2) For audio assessment, following the Audiobox~\cite{vyas2023audiobox} protocol, we employ Audiobox-Aesthetics to determine Perceptual Quality (PQ) and Content Usefulness (CU). To evaluate speech-text alignment, we isolate vocal components via Mel-RoFormer~\cite{wang2024melroformervocalseparationvocal} and compute the Word Error Rate (WER) using Whisper-large-v3~\cite{radford2023robust}. (3) For cross-modal consistency, we utilize CLAP~\cite{elizalde2023clap} for audio-text semantic consistency (TA), VideoCLIP-XL-V2~\cite{wang2024vidprom} for video-text semantic alignment (TV), and ImageBind~\cite{girdhar2023imagebind} for audio-video semantic similarity (AV). Lip-audio synchronization is measured via SyncNet~\cite{Prajwal_2020} (LSE-C and LSE-D) while audio-visual temporal correspondence is captured by the DeSync (DS) score from Synchformer~\cite{iashin2024synchformer} to evaluate absolute time offsets between modal onsets.

\begin{figure*}[t] 
  \vspace{-.1cm}
  \setlength{\abovecaptionskip}{-.1cm}
  \setlength{\belowcaptionskip}{-.4cm}
  \begin{center}     
    \includegraphics[width=\textwidth]{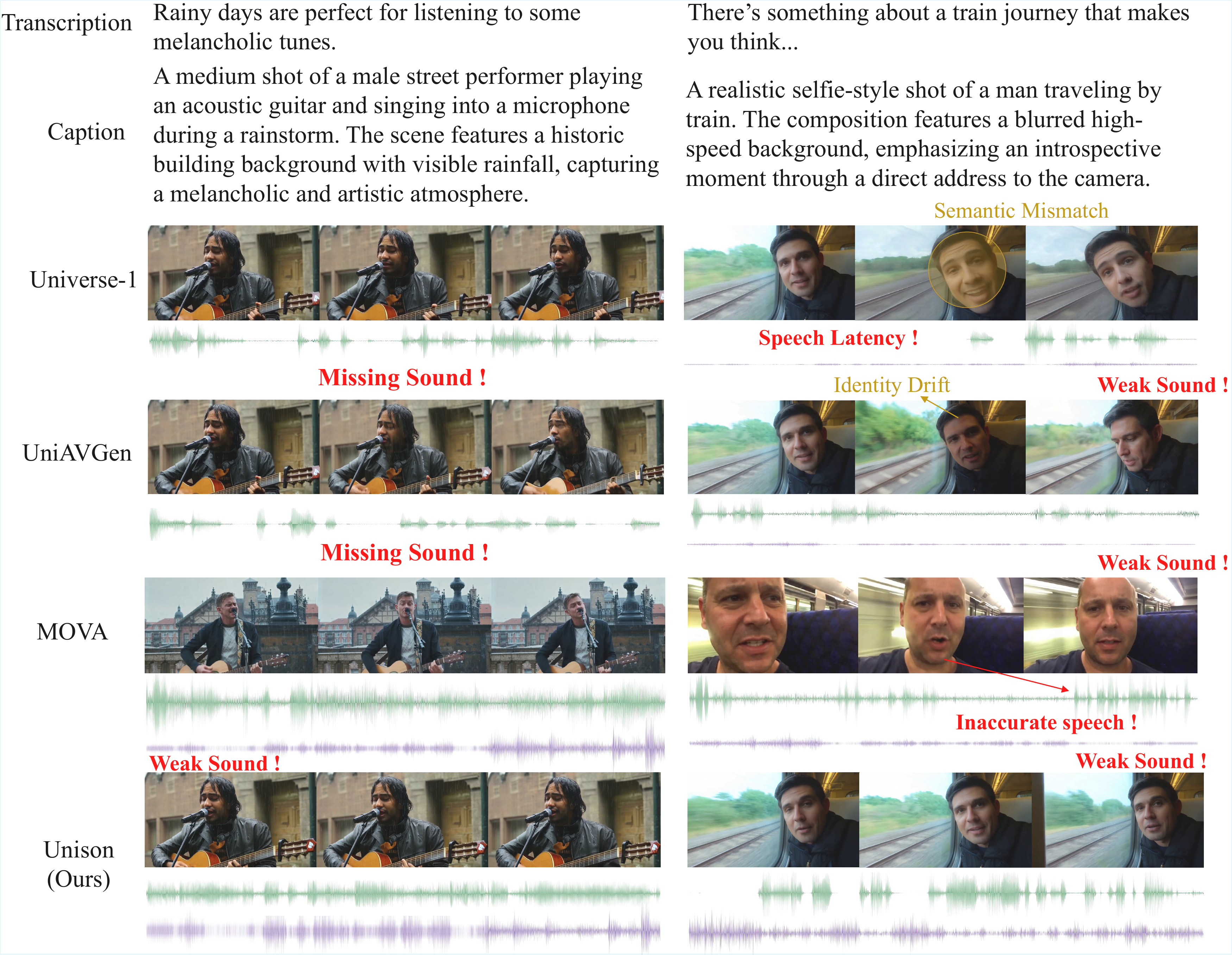}
  \end{center}
  \caption{Qualitative comparison between Unison and the state-of-the-art methods, including Universe-1~\cite{wang2025universe}, UniAVGen~\cite{zhang2025uniavgen} and MOVA~\cite{openmoss_mova_2026}.}
  \label{fig:qual_compare}
\end{figure*}

\begin{figure*}[t]
  \vspace{-.1cm}
  \setlength{\abovecaptionskip}{-.1cm}
  \setlength{\belowcaptionskip}{-.4cm}
  \begin{center}
  \includegraphics[width=\textwidth]{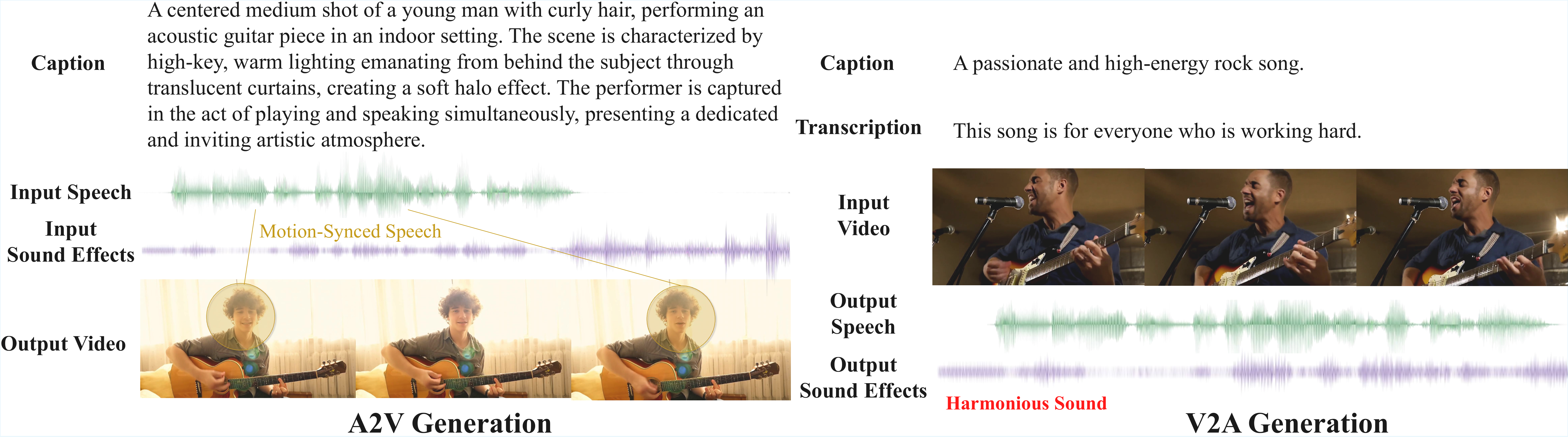}
  \end{center}
  \caption{Bidirectional Synthesis of Audio-to-Video and Video-to-Audio.}
  \label{fig:a2v_v2a}
\end{figure*}
\begin{figure*}[ht]
  \vspace{-.1cm}
  \setlength{\abovecaptionskip}{-.1cm}
  \setlength{\belowcaptionskip}{-.2cm}
  \begin{center}      
    \includegraphics[width=\textwidth]{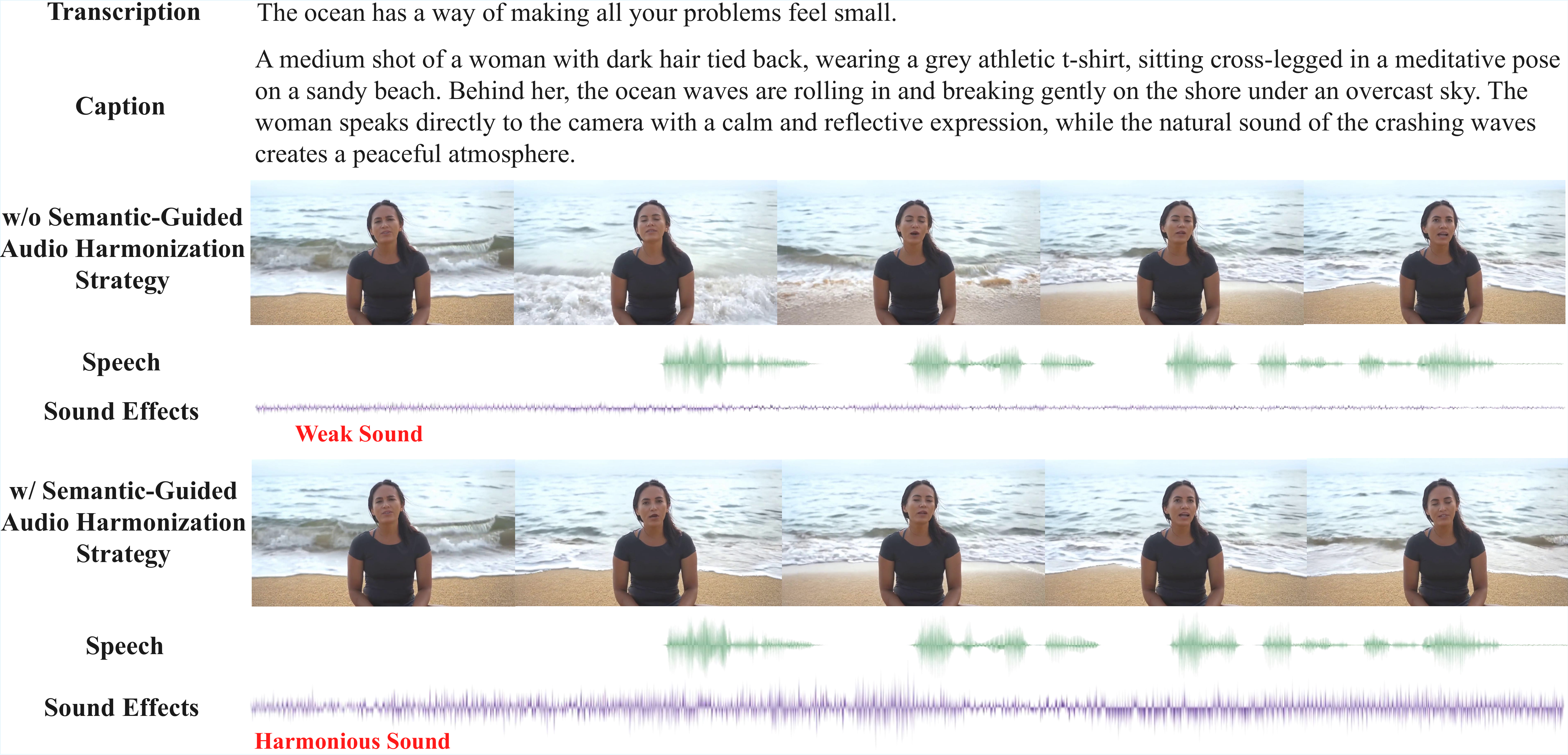}
  \end{center}
  \caption{Ablation experiments on the Semantic-Guided Audio Harmonization Strategy.}
  \label{fig:ablation_audio}
  \begin{center}
      \includegraphics[width=\textwidth]{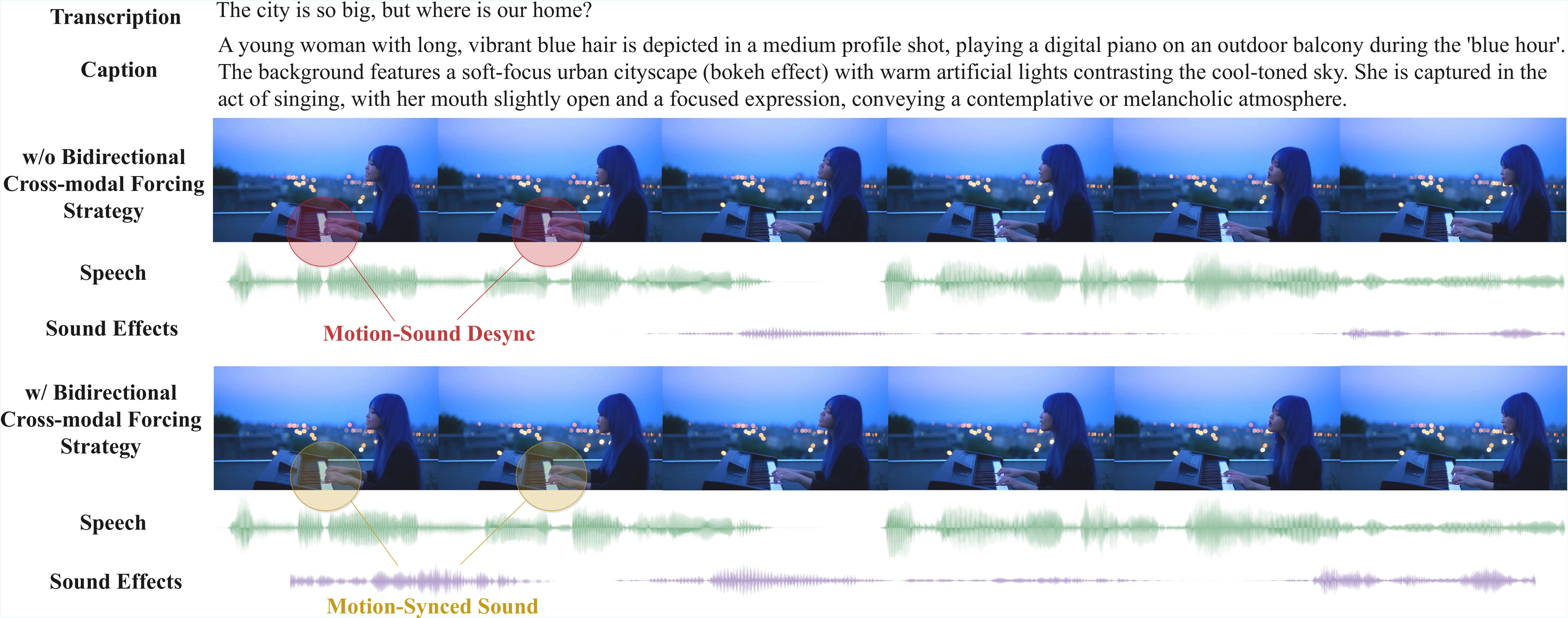}
  \end{center}
  \caption{Ablation experiments on the Bidirectional Cross-modal Forcing Strategy.}
  \label{fig:ablation_video}
\end{figure*}

\subsection{Qualitative Results}
\paragraph{Qualitative Comparisons.}
We evaluate Unison against representative baselines under identical settings. As shown in Fig.~\ref{fig:qual_compare}, Universe-1~\cite{wang2025universe} and UniAVGen~\cite{zhang2025uniavgen} fail to synthesize musical accompaniments and struggle to distinguish singing from standard speech. In locomotive scenes, the aforementioned baselines exhibit severe modal imbalance, with speech volume disproportionately overwhelming engine sounds. MOVA~\cite{openmoss_mova_2026} generates plausible vocals in musical contexts yet suffers from substantial phonetic artifacts in complex environments. Conversely, Unison achieves precise synchronization between motion dynamics and diverse acoustic components, including lip movements and impact transients. Our model maintains superior acoustic layering, ensuring intelligible speech without suppressing salient environmental audio.

\paragraph{Audio-to-Video and Video-to-Audio Generation.} 
Unison leverages decoupled denoising schedules and bidirectional guidance to achieve precise modal translation across both A2V and V2A tasks. In A2V scenarios, audio conditioning aligns motion with salient acoustic features, such as speech and impact transients, while for V2A, the model synthesizes coherent audio streams from visual cues. This capability arises from our progressive cross-modal forcing, which facilitates stable conditioning under heterogeneous denoising dynamics. As shown in Fig.~\ref{fig:a2v_v2a}, Unison delivers high-fidelity synthesis across all translation directions, maintaining robust semantic consistency regardless of the source modality.

\paragraph{Effectiveness of Semantic-Guided Audio Harmonization Strategy.}
We investigate the impact of decoupled audio generation, Bidirectional Audio Cross-Attention (Bi-ACA), and Semantic-Conditioned Gating (SCG) within the audio branch. As shown in the beach scene (Fig.~\ref{fig:ablation_audio}), models lacking these components fail to synthesize harmonious speech and sound-effect representations, leading to speech-dominant audio where ambient waves are significantly attenuated. Conversely, our Semantic-guided Audio Harmonization Strategy enables the adaptive recomposition of structurally distinct speech and sound effects. This mechanism yields sharper acoustic transients and more balanced acoustic mixtures.

\paragraph{Effectiveness of Bidirectional Cross-modal Forcing Strategy.} 
As shown in the piano sequence (Fig.~\ref{fig:ablation_video}), omitting cross-modal forcing leads to temporally drifted accompaniments where audio onsets no longer correspond to specific finger movements. With forcing enabled, the modality at a lower noise level provides an explicit denoising reference for its noisier counterpart, establishing a bidirectional guidance loop that corrects misalignment during generation. The resulting audio exhibits tightly synchronized note onsets and release transients that faithfully follow the observed hand dynamics.

\subsection{Quantitative Results}
\paragraph{Comparison with Baselines.}
We evaluate Unison on a curated test set of 1,000 held-out samples, with ground-truth annotations provided by Gemini to ensure rigorous T2AV and TI2AV assessment. As shown in Table~\ref{tab:comparison}, Unison achieves the highest audio fidelity across all methods, notably leading in PQ, CU, and WER. Despite utilizing a 5B video backbone—nearly 4$\times$ smaller than LTX-2's 19B—Unison exhibits superior cross-modal synchronization. While LTX-2 shows marginal gains in visual texture (VA), Unison remains highly competitive in overall perceptual quality. These results validate that our architectural innovations effectively capture high-fidelity audiovisual correlations without the need for massive parameter scaling.

\begin{table*}[t]
\centering
\caption{Quantitative comparison with state-of-the-art methods. We evaluate performance across three categories: video quality, audio fidelity, and audio-visual synchronization. Best results are in \textbf{bold}, second-best are \underline{underlined}. For more comprehensive and detailed evaluations, please refer to the supplementary material.}
\label{tab:comparison}
\footnotesize
\setlength{\tabcolsep}{1.5pt} 
\resizebox{\linewidth}{!}{%
\begin{tabular}{@{}l l cc ccc cccccc@{}}
\toprule
\multirow{2}{*}{Type} & \multirow{2}{*}{Model} &
\multicolumn{2}{c}{Video Quality} & \multicolumn{3}{c}{Audio Quality} & \multicolumn{6}{c}{Cross-Modal Consistency}\\ 
\cmidrule(lr){3-4} \cmidrule(lr){5-7} \cmidrule(lr){8-13}
& & VA $\uparrow$ & ID $\uparrow$ & PQ $\uparrow$ & CU $\uparrow$ & WER $\downarrow$ & TA $\uparrow$ & TV $\uparrow$ & AV $\uparrow$ & LSE-C $\uparrow$ & LSE-D $\downarrow$ & DS $\downarrow$ \\
\midrule
\multirow{6}{*}{TI2AV} 
& Universe-1~\cite{wang2025universe} & 3.77 & 4.42 & 5.95 & 5.21 & 0.52 & 3.37 & 25.57 & 0.62 & 2.32 & -- & 0.50 \\
& Ovi~\cite{low2025ovi} & 3.94 & 4.42 & 6.25 & 5.51 & 0.43 & 3.48 & 25.86 & 0.87 & 2.81 & 9.12 & 0.12 \\
& UniAVGen~\cite{zhang2025uniavgen} & 4.02 & 4.46 & 6.18 & 5.48 & 0.33 & 3.42 & 25.99 & 0.81 & 2.89 & 9.49 & 0.15 \\
& MOVA~\cite{openmossteam2026movascalablesynchronizedvideoaudio} & 4.01 & 4.52 & 6.28 & 5.52 & 0.29 & 3.58 & 25.97 & 0.88 & 3.24 & 7.92 & 0.13 \\
& LTX-2~\cite{hacohen2026ltx2efficientjointaudiovisual} & \textbf{4.15} & \textbf{4.61} & \underline{6.30} & \underline{5.58} & \underline{0.25} & \textbf{3.65} & \textbf{26.24} & \underline{0.89} & \textbf{3.45} & \textbf{7.62} & \underline{0.10} \\
& Unison (Ours) & \underline{4.02} & \underline{4.53} & \textbf{6.34} & \textbf{5.61} & \textbf{0.22} & \underline{3.61} & \underline{26.17} & \textbf{0.91} & \underline{3.30} & \underline{7.88} & \textbf{0.08} \\
\midrule
\multirow{5}{*}{T2AV} 
& JavisDiT~\cite{liu2025javisdit} & 3.29 & 4.52 & 4.83 & 3.73 & 1.81 & 3.53 & 24.31 & 0.49 & 1.81 & -- & 0.53 \\
& Ovi~\cite{low2025ovi} & 4.22 & 4.51 & 6.08 & 5.65 & 0.18 & 3.55 & 25.99 & \underline{0.83} & 3.47 & 8.05 & 0.08 \\
& LTX-2~\cite{hacohen2026ltx2efficientjointaudiovisual} & \textbf{4.63} & \textbf{4.68} & \underline{6.12} & \underline{5.72} & \underline{0.11} & \textbf{3.74} & \textbf{26.35} & 0.81 & \textbf{3.62} & \textbf{7.75} & \underline{0.07} \\
& Unison (Ours) & \underline{4.51} & \underline{4.59} & \textbf{6.17} & \textbf{5.78} & \textbf{0.09} & \underline{3.62} & \underline{26.21} & \textbf{0.86} & \underline{3.55} & \underline{7.95} & \textbf{0.06} \\
\bottomrule
\end{tabular}
}%
\end{table*}

\begin{figure*}[t]
  \vspace{-.1cm}
  \setlength{\abovecaptionskip}{-.1cm}
  \setlength{\belowcaptionskip}{-.4cm}
  \begin{center}  
    \includegraphics[width=\textwidth]{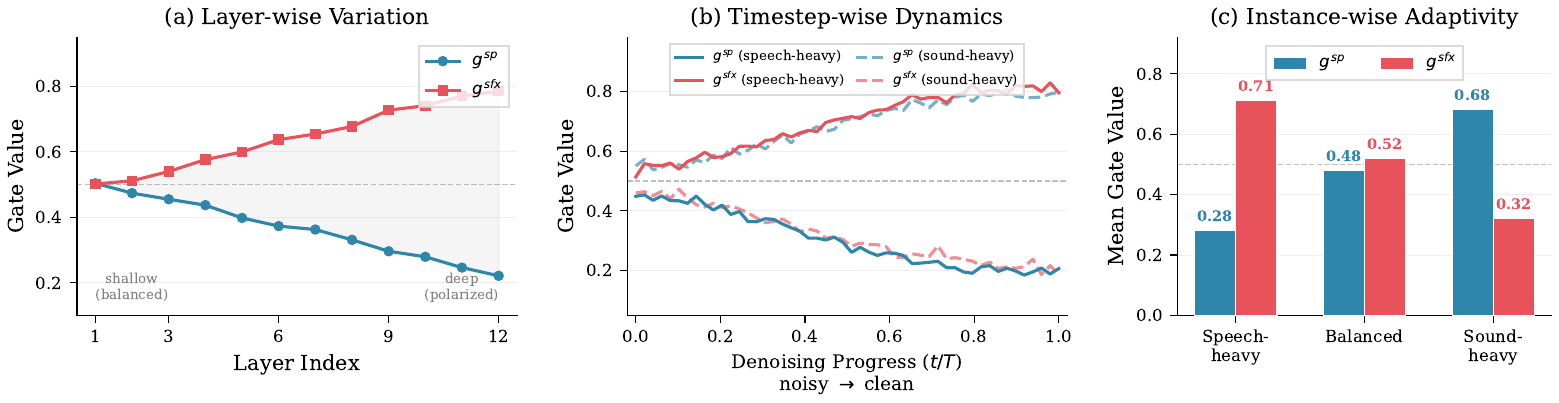}
  \end{center}
  \caption{Analysis of SCG gate behavior. (a) \textbf{Layer-wise}: gate polarization increases with model depth. (b) \textbf{Timestep-wise}: gate divergence intensifies as denoising progresses. (c) \textbf{Instance-wise}: mean gate values across semantic categories, demonstrating content-adaptive modulation.}
  \label{fig:gate_analysis}
\end{figure*}

\paragraph{Analysis of SCG Gating Behavior.} 
To verify that SCG learns dynamic, content-aware modulation, we visualize the gate values $g^{sp}$ and $g^{sfx}$ across different dimensions (Fig.~\ref{fig:gate_analysis}). 
\textit{Layer-wise}: Shallow layers maintain balanced gates ($g \approx 0.5$) for coarse structure formation, while deeper layers exhibit increasing polarization for semantic refinement. 
\textit{Timestep-wise}: Gate divergence intensifies as denoising progresses; $g^{sp}$ and $g^{sfx}$ stay moderate at high noise levels but diverge significantly as content crystallizes, ensuring interference suppression. 
\textit{Instance-wise}: SCG mitigates the dominance of speech over subtle environmental textures via dynamic rebalancing. In sports broadcasting, the mechanism constrains the narration stream ($g^{sp}=0.62$) to prevent it from masking critical acoustic cues. This safeguards high-frequency transients of the stadium atmosphere ($g^{sfx}=0.38$), ensuring that impact sounds and crowd cheers remain perceptible despite high-volume vocal inputs.

\paragraph{Ablation Study on Components.} 
Table~\ref{tab:ablation_arch} evaluates four core modules: SGHS (Semantic-Guided Harmonization Strategy), Bi-ACA (Bidirectional Audio Cross-Attention), SCG (Semantic-Conditioned Gating), and CMFS (Cross-Modal Forcing Strategy). We report video quality (VA), audio fidelity (PQ), and synchronization (LSE-C, DS). Among audio-side components, removing SGHS causes the most significant PQ degradation ($6.12$ vs.\ $6.34$) due to spectral interference, and eliminating Bi-ACA and SCG further reduces PQ and lip-sync precision by weakening bidirectional context and text-driven suppression. Notably, these audio-specific ablations leave VA largely unaffected ($3.99$--$4.01$), confirming their modularity. In contrast, removing CMFS reveals a distinct degradation pattern: DS rises sharply to $0.19$ and LSE-C drops to $3.02$, the poorest alignment scores across all settings. Furthermore, VA decreases to $3.91$, indicating the video branch benefits from audio-stream guidance to reinforce visual coherence. These results confirm a functional complementarity: audio-side modules ensure acoustic fidelity, and CMFS establishes robust audio-visual synchronization.

\begin{table}[t]
\centering
\footnotesize
% 第一个表格：左半部分
\adjustbox{valign=t}{%
\begin{minipage}{0.48\textwidth}
\centering
\caption{Ablation study on the core components of Unison.}
\label{tab:ablation_arch}
\begin{tabular}{lcccc}
\toprule
Settings & VA $\uparrow$ & PQ $\uparrow$ & LSE-C $\uparrow$ & DS $\downarrow$ \\
\midrule
w/o SGHS  & 3.99 & 6.12 & 3.08 & 0.15 \\
w/o Bi-ACA & 4.00 & 6.20 & 3.18 & 0.11 \\
w/o SCG   & 4.01 & 6.21 & 3.22 & 0.10 \\
w/o CMFS  & 3.91 & 6.24 & 3.02 & 0.19 \\
Ours      & \textbf{4.02} & \textbf{6.34} & \textbf{3.30} & \textbf{0.08} \\
\bottomrule
\end{tabular}
\end{minipage}
}
\hfill
% 第二个表格：右半部分
\adjustbox{valign=t}{%
\begin{minipage}{0.48\textwidth}
\centering
\caption{Ablation study on the training strategies of Unison.}
\label{tab:ablation_training}
\begin{tabular}{lcccc}
\toprule
Settings & VA $\uparrow$ & PQ $\uparrow$ & LSE-C $\uparrow$ & DS $\downarrow$ \\
\midrule
SyncOnly     & 3.90 & 6.10 & 3.12 & 0.17 \\
IndepOnly    & 3.95 & 6.18 & 3.28 & 0.14 \\
PF(Ours) & \textbf{4.02} & \textbf{6.34} & \textbf{3.30} & \textbf{0.08} \\
\bottomrule
\end{tabular}
\end{minipage}
}
\end{table}

\paragraph{Ablation of Training Strategy.} 
Table~\ref{tab:ablation_training} evaluates three training schedules for cross-modal forcing. The SyncOnly baseline, enforcing $t_v{=}t_a$ throughout, yields the lowest performance (VA 3.90, DS 0.17), confirming that uniform noise prevents the model from exploiting denoising asymmetries. IndepOnly, which applies fully decoupled timesteps without a warmup, improves alignment (DS 0.14) but destabilizes optimization, capping LSE-C at 3.28 due to abrupt noise gaps. In contrast, our ProgForcing (PF) schedule implements a three-phase curriculum—synchronous warmup, incremental decoupling, and full independence—to progressively introduce challenging noise configurations. This strategy achieves superior results across all metrics (VA 4.02, PQ 6.34, LSE-C 3.30, DS 0.08), demonstrating that a gradual transition is essential for both stable optimization and precise temporal alignment.

\begin{table}[htbp]
  \centering
  \caption{Results of the user study. 40 samples $\times$ 25 participants (1{,}000 mean-rank votes; lower is better).}
  \label{fig:user_study}
  \begin{tabular}{lcccc}
    \toprule
    Method & Lip-sync$\downarrow$ & \shortstack{Speech-Sound\\Harmony$\downarrow$} & \shortstack{Motion-Audio\\Align.$\downarrow$} & \shortstack{Gemini\\Score$\downarrow$} \\
    \midrule
    UniAVGen & 3.51 & 3.45 & 3.42 & 3.48 \\
    MOVA & 2.58 & 2.77 & 2.89 & 2.48 \\
    LTX-2 & \textbf{1.74} & \underline{1.95} & \textbf{1.89} & \underline{2.05} \\
    Unison & \underline{1.86} & \textbf{1.55} & \underline{1.92} & \textbf{1.68} \\
    \bottomrule
  \end{tabular}
\end{table}

\paragraph{User Study.}
We conducted a user study with 40 samples $\times$ 25 participants from diverse backgrounds (1{,}000 mean-rank votes; lower is better), evaluating lip-speech synchrony, speech-sound harmony, and motion-audio alignment (considering both speech and environmental sounds). We further incorporate an additional evaluation column by scoring Motion-Speech-SFX coherence via Gemini. Participants were required to rank shuffled videos across different methods, including UniAVGen~\cite{zhang2025uniavgen}, MOVA~\cite{openmoss_mova_2026}, and LTX-2~\cite{hacohen2026ltx2efficientjointaudiovisual}. As shown in Table~\ref{fig:user_study}, our approach achieved the highest scores in both speech-sound harmony and motion-audio alignment, as well as the best overall Gemini coherence score. While our lip-speech synchrony performance was second only to LTX-2, our method received the dominant overall preference across the integrated metrics, demonstrating superior holistic audio-visual quality.

\section{Conclusion.}
In this work, we present Unison, a framework that achieves superior motion-speech-sound synchronization. By implementing decoupled acoustic modeling alongside a progressive cross-modal forcing strategy, Unison effectively resolves the temporal misalignments and modal interference typical of joint synthesis. Extensive evaluations demonstrate that Unison achieves state-of-the-art performance in cross-modal consistency and audio perceptual fidelity. 

\section*{Acknowledgements}
This work was supported by the NSFC Regional Innovation and Development Joint Fund under Grant U25A20537, and the National Key Research and Development Program of China under Grant 2024YFC3015600.

\clearpage  % TODO FINAL: This \clearpage needs to be removed from both review and camera-ready versions.

% ---- Bibliography ----
%
% BibTeX users should specify bibliography style 'splncs04'.
% References will then be sorted and formatted in the correct style.
%
\bibliographystyle{splncs04}  % 按引用出现顺序排序；投稿时如需 splncs04 可改回
\bibliography{main}
\end{document}

% ---------------------------------------------------------------
% 标题部分
% ---------------------------------------------------------------
\title{Appendix} 

\titlerunning{Unison}

\author{Shihao Cheng\inst{1,4\dagger} \and
Jiaxu Zhang\inst{2\dagger} \and
Quanyue Song\inst{3,4} \and
Shansong Liu\inst{4\ddagger} \and
Zhizhi Guo\inst{4} \and
Xiao-Lei Zhang\inst{5} \and
Chi Zhang\inst{5} \and
Xuelong Li\inst{5} \and
Zhigang Tu\inst{1}\textsuperscript{\Letter}}
\authorrunning{S.~Cheng et al.}
\institute{State Key Laboratory of Information Engineering in Surveying, Mapping and Remote Sensing, Wuhan University, Wuhan, China \and
ByteDance, China \and
State Key Laboratory of Human-Machine Hybrid Augmented Intelligence, Institute of Artificial Intelligence and Robotics, Xi'an Jiaotong University, China \and
China Telecom Artificial Intelligence Technology (Beijing) Co., Ltd., China \and
Institute of Artificial Intelligence (TeleAI), China Telecom, China
}

\maketitle

\begingroup
\renewcommand\thefootnote{}\footnotetext{$^{\dagger}$ Equal contribution.}
\renewcommand\thefootnote{}\footnotetext{$^{\ddagger}$ Project leader.}
\renewcommand\thefootnote{}\footnotetext{\textsuperscript{\Letter} Corresponding author.}
\endgroup

\section{Overview}
\label{sec:overview}
This supplementary material begins by detailing the multi-stage data filtering and processing pipeline in Sec.~\ref{sec:data}. Subsequently, Sec.~\ref{sec:bench} describes the composition and scenario-specific categorization of the curated evaluation benchmark, which consists of 1,000 prompts and corresponding transcriptions. To provide deeper insights into the internal mechanisms of the model, an extended analysis of visualization ablations is presented in Sec.~\ref{sec:vis_ablation}, followed by a thorough hyperparameter study in Sec.~\ref{sec:hyperparam} that covers aspects such as progressive training schedules. Finally, Sec.~\ref{sec:more_examples} offers a diverse gallery of additional qualitative examples, showcasing the performance of Unison in various human-centric scenarios.

% ---------------------------------------------------------------
% 1) 数据筛选处理流程
% ---------------------------------------------------------------
\section{Data Filtering and Processing Pipeline}
\label{sec:data}
To ensure robust temporal semantic alignment and high-fidelity joint audio-video generation, we implement a multi-stage data filtering and processing pipeline. This pipeline integrates assessments of video quality, audio quality, and synchronization accuracy.

In the video quality assessment stage, the system first filters videos with aspect ratios exceeding $2$ or falling below $0.5$ to ensure format compatibility. Subsequently, the pipeline removes logos and subtitles through a fixed-cropping strategy and evaluates the visual quality of the remaining content via an image aesthetics predictor. Notably, a shot transition detector is employed to identify hard cuts and abrupt scene changes. The pipeline ultimately retains only those video segments characterized by high aesthetic scores and at most one shot transition, a criterion designed to maintain visual consistency and effectively minimize potential temporal discontinuities.

Audio quality assessment involves calculating audio energy statistics to exclude segments that are either nearly silent or characterized by excessive background noise. Subsequently, the Audiobox Aesthetics (AES) model~\cite{vyas2023audioboxunifiedaudiogeneration} is employed to quantify production quality (PQ) and content enjoyment (CE). By establishing multi-dimensional scoring thresholds, this process filters out samples with missing, corrupted, or low signal-to-noise ratio (SNR) audio, while also removing segments with suboptimal acoustic performance. Such filtering enhances the overall quality and aesthetic coherence of the training data. Furthermore, Mel-Roformer~\cite{wang2024melroformervocalseparationvocal} is utilized for source separation, followed by automatic speech recognition (ASR) to identify and select segments with complete audio content. Subsequently, a sliding window approach is adopted to screen fixed-length audio segments that satisfy predefined integrity requirements, thereby increasing the volume of high-quality trainable samples.

To guarantee precise temporal correspondence between audio and video streams, the framework utilizes two complementary synchronization evaluators to capture both local and global dependencies. Specifically, SyncNet~\cite{Chung16a} assesses fine-grained lip-sync scores, while Synchformer~\cite{iashin2024synchformerefficientsynchronizationsparse} measures broader global audio-visual alignment. Consequently, segments that do not meet the following criteria are discarded:
\begin{equation}
LSE-C > 2.5, \quad LSE-D < 9, \quad \text{DeSync} < 0.1
\end{equation}
Notably, this dual-scale filtering step effectively eliminates samples characterized by significant lip-sync errors or broader temporal misalignments. By enforcing these stringent constraints, we ensure high-quality synchronization and reliable cross-modal supervision during the training phase.

Finally, the joint audio-video captioning pipeline extracts fine-grained, temporally ordered semantic descriptions. Through structured prompting, Gemini performs joint analysis and fused annotation of audio-visual signals. For each sample, the model generates temporal annotations describing visual and auditory content, including the background environment, ambient sound and music, character appearance, actions, and facial expressions, as well as spoken content, stylistic emotional tone, and camera motion patterns. Beyond these core attributes, the pipeline integrates additional task-specific annotation fields to ensure comprehensive coverage of audio-visual semantics. The prompts are designed to enforce strict temporal awareness, enabling the precise extraction of information across different timestamps. This automated labeling process provides high-quality, temporally grounded semantic supervision to guide the learning of joint audio-video generation and temporal alignment.

% ---------------------------------------------------------------
% 2) 可视化消融实验的分析
% ------------------------------------------------------------
\section{Evaluation Benchmark}
\label{sec:bench}
To evaluate the performance of Unison comprehensively in comparison with existing models, we curate a benchmark comprising 1,000 evaluation prompts and the corresponding transcriptions. This benchmark encompasses three representative scenarios: talking head, ambient sound, and temporally complex scenes. All test samples remain strictly unseen during the training phase to ensure a rigorous assessment of generalization capabilities.

The talking head subset, consisting of 400 entries, features speakers depicted from diverse perspectives. Beyond standard lip-synchronization, this category specifically evaluates the accuracy of speech generation and the rhythmic consistency between facial kinetics and acoustic onsets. Such metrics directly reflect the efficacy of the bidirectional cross-modal forcing paradigm in establishing stable temporal coordination.

The ambient sound subset, comprising 200 entries, includes environmental sounds and background music descriptions. This subset is designed to assess audio-visual alignment and, crucially, the acoustic richness and balance of the generated output. We examine whether the semantic-guided harmonization effectively prevents speech components from overshadowing subtle contextual sounds, thereby maintaining perceptual realism.

The temporally complex subset, comprising 400 entries, depicts scenes with clearly ordered actions—such as singing while playing an instrument—that intertwine speech and ambient sounds. This category evaluates temporal semantic alignment and the mitigation of intra-audio interference. The focus remains on whether the model can disentangle heterogeneous acoustic elements while ensuring that motion, speech, and sound evolve in unison to yield synchronized and perceptually balanced results.

Furthermore, for all entries containing speech, the transcriptions are systematically adjusted to ensure complete semantic sequences, thereby providing a consistent basis for the rigorous evaluation of long-range temporal coherence.

\section{Visualization Ablation Analysis}
\label{sec:vis_ablation}
We provide an extended analysis of the qualitative ablation results presented in the main paper, with a specific focus on the removal of components within the audio branch.

\begin{figure}[t]
  \centering
  \includegraphics[width=\textwidth]{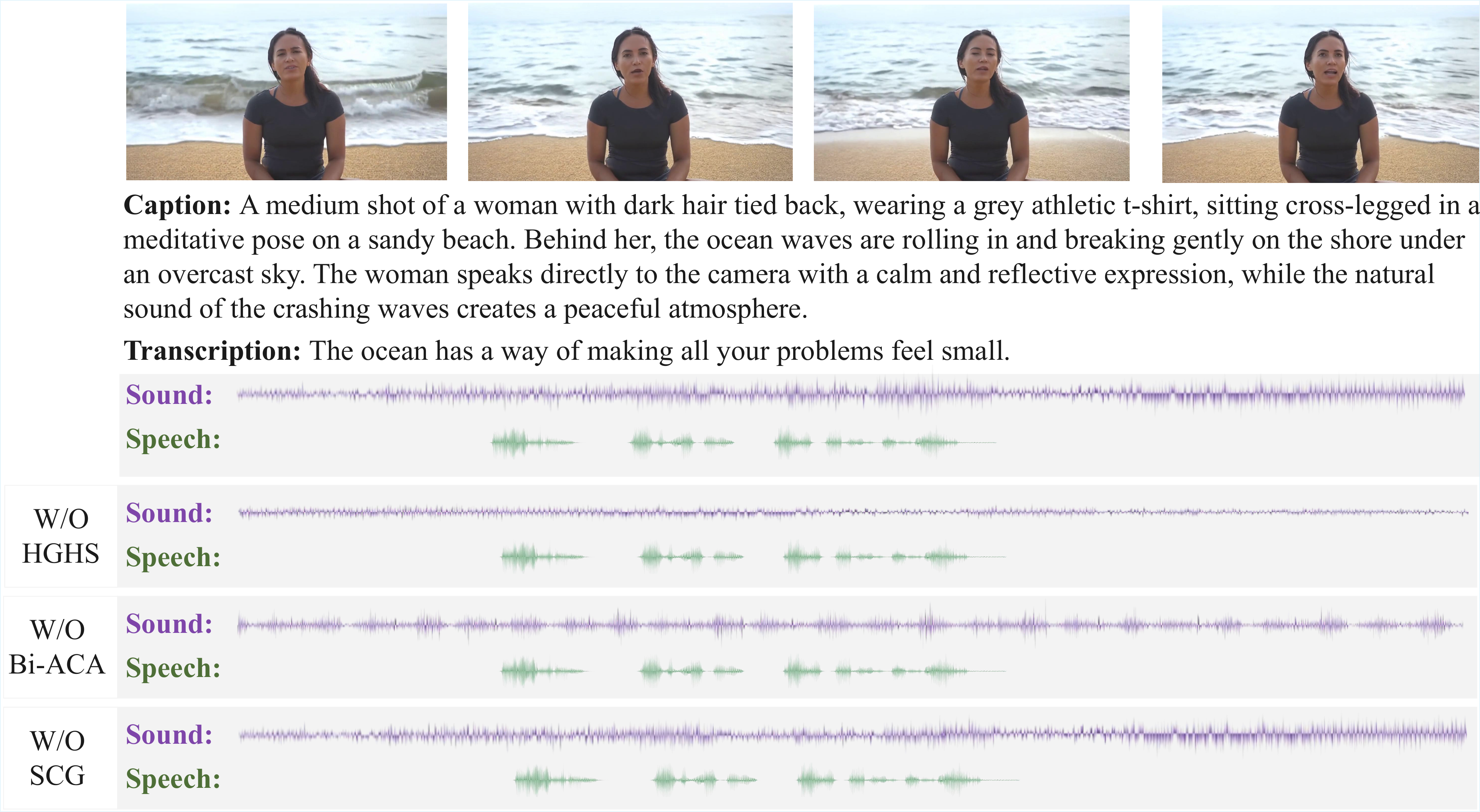}
  \caption{Visual Analysis of Audio Branch Ablations.}
  \label{fig:vis_ablation}
\end{figure}

\paragraph{Spectrogram Structure Reveals Hierarchical Degradation Under Audio-Side Ablations}
Figure~\ref{fig:vis_ablation} presents an extended visualization of the beach scene spectrogram analyzed in the main text. The removal of HGHS leads to a single-stream representation in which speech formants overlap with the broadband energy of waves across the entire temporal axis. The harmonics-to-noise ratio shifts toward the speech components, which is consistent with the subjective perceptual dominance observed in the quantitative ablation study, where the PQ decreases from 6.34 to 6.12. While removing Bi-ACA alone maintains the dual-stream structure, the frequency bands of the sound effects lose temporal coherence with the speech envelope. Under this condition, when speech and sound effects undergo cross-attention with the video independently, the absence of interaction causes these components to be processed as misaligned audio-visual noise, thereby creating a sensory impression of two layers generated in isolation. The removal of SCG results in a collapse of gating values into a neutral range, which leads to blurred boundaries between vocal formants and ambient transients in the final output. In contrast, the full model achieves a distinct hierarchical separation: speech occupies well-defined formant regions, while ambient waves form mid-to-low frequency bands with sharp onset transients, consistent with the semantic description of a speaker situated against a coastal background.

\paragraph{Visual Comparison Highlights the Complementarity of Audio and Video Modules}
By juxtaposing audio-side ablations (without HGHS, Bi-ACA, and SCG) with the bidirectional cross-modal ablation (without CMFS) discussed in the main text, a distinct functional division is observed. The removal of audio-side components primarily degrades spectral and phonetic quality, with PQ and WER being the most significantly affected metrics, while lip-sync scores (LSE-C) and global alignment (DS) remain relatively stable. In contrast, the removal of CMFS results in the most substantial decline in LSE-C (3.02) and DS (0.19), even though the spectrograms remain reasonably structured; thus, the degradation is temporal rather than spectral. This complementarity validates the modular design: semantic-guided harmonization ensures acoustic fidelity, whereas cross-modal forcing enforces temporal alignment. 

\section{Hyperparameter Analysis}
\label{sec:hyperparam}
We conduct a systematic ablation study on the key hyperparameters governing bidirectional cross-modal forcing strategies to empirically verify the efficacy of our specific parameter configurations. Furthermore, this analysis aims to elucidate how these hyperparameters modulate the trade-off between cross-modal alignment and generative stability within the proposed framework.

\paragraph{Timestep Gap Constraint $\Delta_{\max}$ Mediates Guidance Strength and Stability}The constraint $|t_v - t_a| \leq \Delta_{\max}$ limits the maximum noise disparity between modalities. At $\Delta_{\max}=0.15$, the DS increases to 0.10 and the PQ reaches 6.32; this tight bound restricts the range over which the cleaner modality can guide the noisier one, thereby weakening the intended cross-modal supervisory signal. Conversely, at $\Delta_{\max}=0.35$, the DS degrades to 0.11 and the PQ to 6.26, while training loss curves exhibit higher variance. The enlarged gap amplifies stochastic oscillations in the leading direction $d$, which destabilizes the bidirectional guidance loop. $\Delta_{\max}=0.25$ yields a DS of 0.08 and a PQ of 6.34, representing an optimal balance where sufficient asymmetry enables effective forcing without sacrificing optimization stability. This result validates the design rationale: moderate decoupling is essential for cross-modal guidance to be effective, whereas excessive decoupling undermines the learning dynamics.

\paragraph{Direction-Aware Loss Weight $\lambda$ Modulates Cross-Modal Supervision Emphasis}
The hyperparameter $\lambda$ in $w_v = 1 + \lambda \cdot d$ and $w_a = 1 + \lambda \cdot (1-d)$ governs the intensity with which the noisier modality is incentivized to learn from its cleaner counterpart. At $\lambda=0.3$, the model yields a DS of 0.09, a PQ of 6.32, and an LSE-C of 3.22, suggesting that a lower weight insufficiently enforces cross-modal conditioning. Conversely, at $\lambda=0.7$, while the DS remains stable at 0.08, the PQ declines to 6.18 and the VA to 3.97; notably, this aggressive reweighting overly biases the optimization toward alignment at the expense of per-branch fidelity, particularly for the video stream. Our proposed configuration of $\lambda=0.5$ achieves optimal performance with a DS of 0.08, a PQ of 6.34, an LSE-C of 3.30, and a VA of 4.02, thereby effectively balancing alignment gains with unimodal quality. These empirical results demonstrate that an equal weighting of the base loss and the direction-aware component provides a robust equilibrium for the bidirectional forcing objective.

\begin{table}[t]
  \centering
  \caption{Hyperparameter ablation on DS, PQ, LSE-C, and VA.}
  \label{tab:hyperparam}
  \footnotesize
  \begin{tabular}{llcccc}
    \toprule
    Hyperparameter & Setting & DS $\downarrow$ & PQ $\uparrow$ & LSE-C $\uparrow$ & VA $\uparrow$ \\
    \midrule
    $\Delta_{\max}$ & 0.15 & 0.10 & 6.32 & 3.22 & 4.01 \\
    & 0.35 & 0.11 & 6.26 & 3.15 & 3.96 \\
    & 0.25 (Ours) & 0.08 & 6.34 & 3.30 & 4.02 \\
    \midrule
    $\lambda$ & 0.3 & 0.09 & 6.32 & 3.22 & 4.01 \\
    & 0.7 & 0.08 & 6.18 & 3.28 & 3.97 \\
    & 0.5 (Ours) & 0.08 & 6.34 & 3.30 & 4.02 \\
    \bottomrule
  \end{tabular}
\end{table}

\section{Additional Qualitative Examples}
\label{sec:more_examples}
We present an extensive gallery of qualitative results encompassing a diverse range of human-centric scenarios. To facilitate a comprehensive evaluation, the selection incorporates challenging cases characterized by complex temporal dynamics, varied environmental contexts, and fine-grained facial expressions. These examples further substantiate the robustness and versatility of the proposed framework in synthesizing high-fidelity and precisely synchronized audio-visual content across disparate domains.

% 第一张图
\begin{figure}[t]
  \centering
  \includegraphics[width=\textwidth]{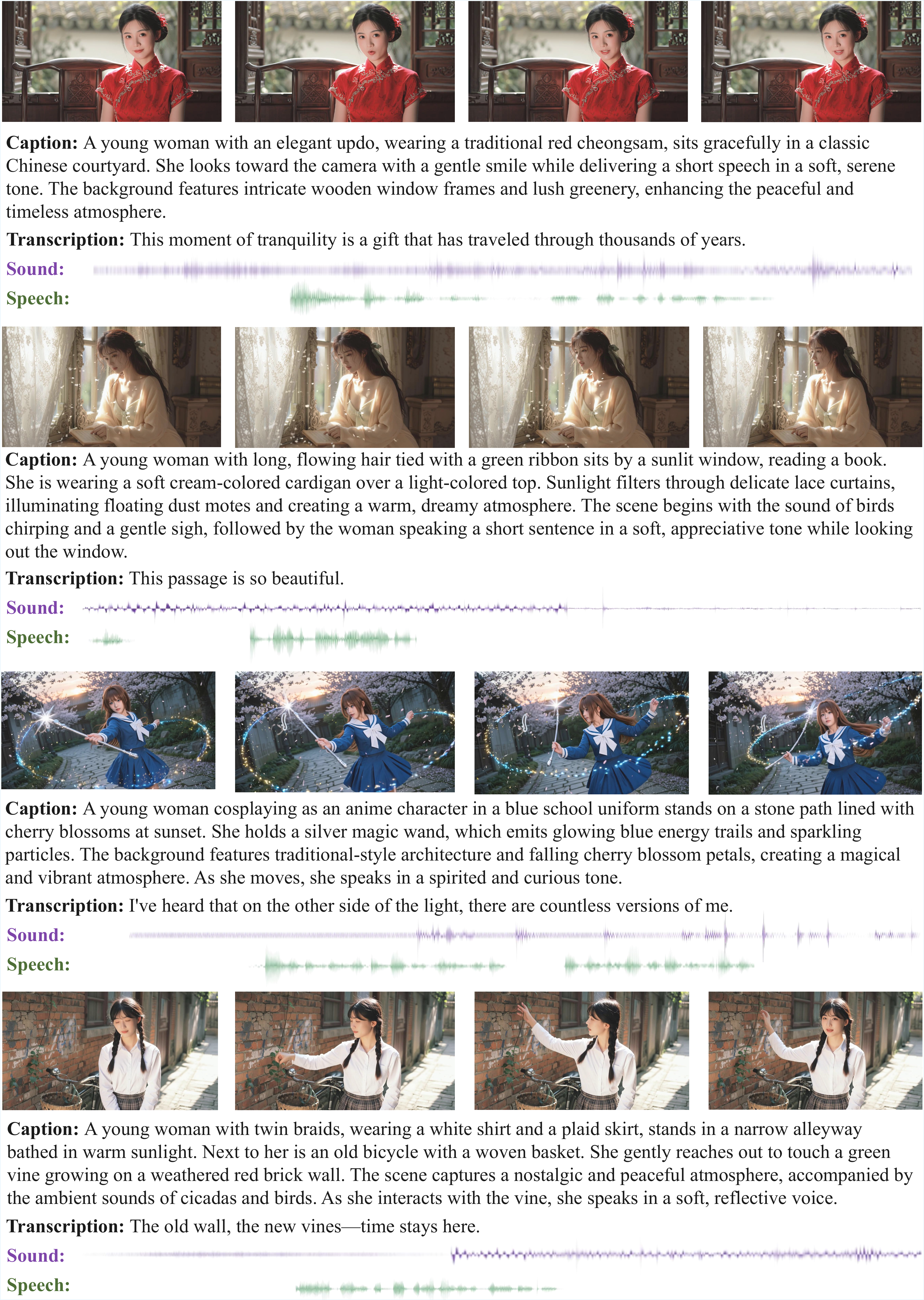}
  \caption{Additional Visual Examples across Diverse Scenarios (Part I).}
  \label{fig:more_examples_1}
\end{figure}

% 第二张图
\begin{figure}[t]
  \centering
  \includegraphics[width=\textwidth]{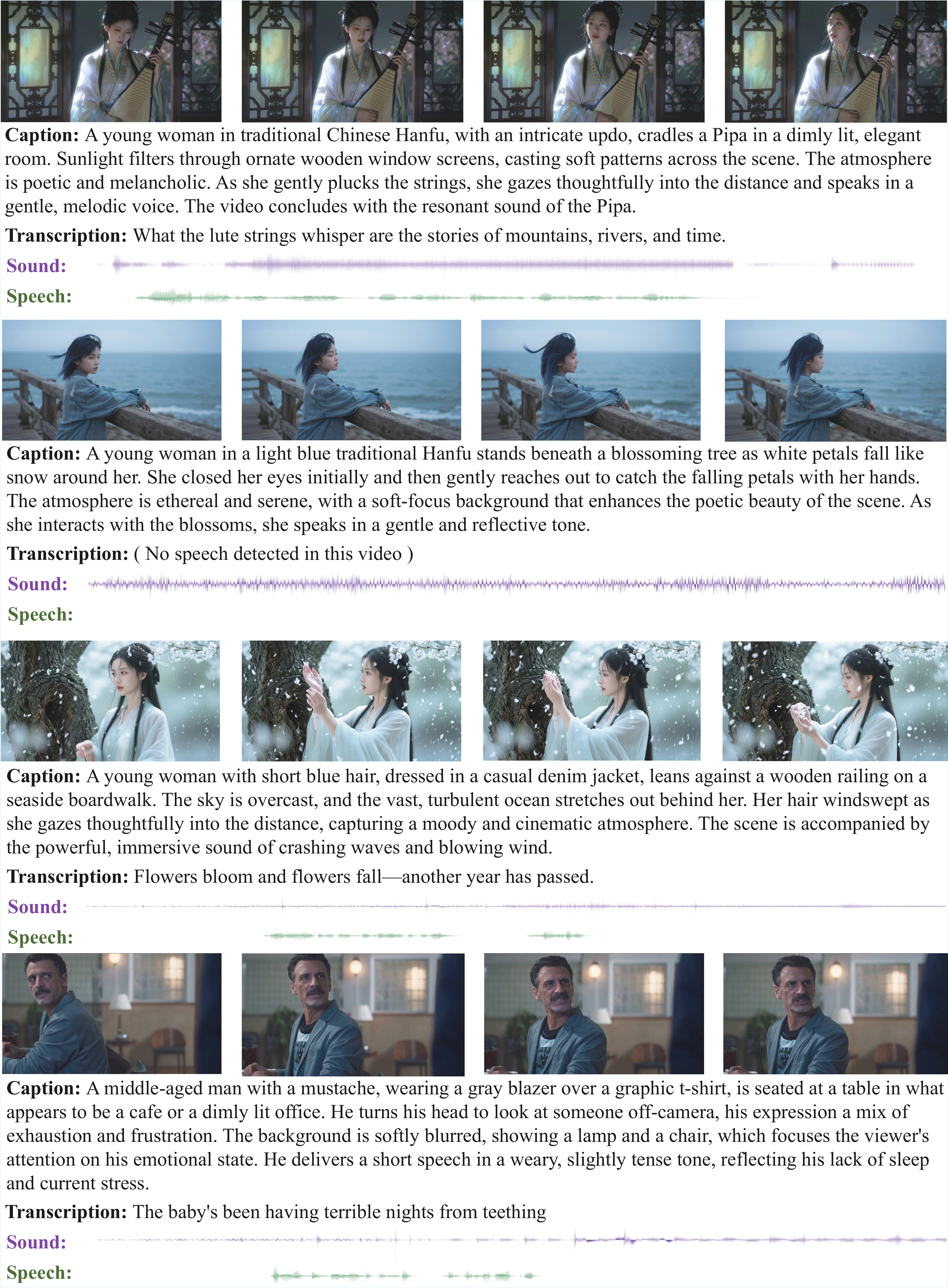}
  \caption{Additional Visual Examples across Diverse Scenarios (Part II).}
  \label{fig:more_examples_2}
\end{figure}
\clearpage
% ---------------------------------------------------------------
% 参考文献
% ---------------------------------------------------------------
\bibliographystyle{unsrt}
\bibliography{main}